\documentclass[]{article}
\usepackage{graphicx}
\usepackage{subfigure}
\usepackage{hyperref}
\hypersetup{pdftitle={Radar shadow detection in SAR images using DEM and projections}, colorlinks=true, citecolor=blue, linkcolor=red}
\pdfoutput=1
\begin{document}

\title{Radar shadow detection in SAR images using DEM and projections}

\author{V. B. S. Prasath\thanks{Department of Computer Science,
University of Missouri-Columbia, MO 65211 USA. E-mail: prasaths@missouri.edu. This work was done while the first author was visiting IPAM, University of California Los Angeles, CA, USA.}\and O. Haddad}
\date{}
\maketitle

\begin{abstract}
\noindent Synthetic aperture radar (SAR) images are widely used in target recognition tasks nowadays. In this letter, we propose an automatic approach for radar shadow detection and extraction from SAR images utilizing geometric projections along with the digital elevation model (DEM) which corresponds to the given geo-referenced SAR image. First, the DEM is rotated into the radar geometry so that each row would match that of a radar line of sight. Next, we extract the shadow regions by processing row by row until the image is covered fully. We test the proposed shadow detection approach on different DEMs and a simulated 1D signals and 2D hills and volleys modeled by various variance based Gaussian functions. Experimental results indicate the proposed algorithm produces good results in detecting shadows in SAR images with high resolution.

\end{abstract}

\section{Introduction}\label{intro}

Synthetic aperture radar (SAR) contains a large amount of information that offers excellent performance compared to other satellite imaging systems. Despite the advantages SAR images have shortcomings such as relief effect due to the use of imaging radar which illuminates the ground from oblique views~\cite{Raney}. Thus, the image produced by such radars contain spatial distortions related to geometric characteristics which are in inherent in the formed acquisition geometry~\cite{Raney,Mikhail}. Some of these geometric distortions can be reversed, for example in the case of ``foreshortening"~\cite{Raney}, if the terrain model is available a priori, it can be removed using the geocoding process. 

There have been efforts to recover or identify shadows in real and very high resolution images~\cite{TongLinHRshadows13,AdelineVHRshadow13}, and for shape from shadow~\cite{AbramsSfShadow13}, moving cast shadows~\cite{SaninShadowSurveyPR12}, determining building heights~\cite{ComberHRshadow12} applications. Unfortunately, in the case of shadows present in a SAR such remedies are not possible due to irreversible lose of information. In fact, the radar will not be able to recover hidden areas which are result of lack of information concerning those particular spatial regions. Shadow regions in SAR imagery are unique in the sense that they can be characterized by the absence of distortion as well as the absence of typical speckle noise which affects other parts of the images. Nevertheless, a dark shadow area present in a SAR image can either depict a smooth region corresponding to flat or bare soil or water surfaces. This induces a confusion in inferring these regions~\cite{Jahangir,Papson}. Moreover, typically the differences between these areas are huge with same radar image representation. To detect shadows in SAR images, previous works rely on the characteristic of shadow areas such as the absence of speckles~\cite{Mikhail} or distortion free areas. These were complicated and time consuming procedures since detecting all the areas not affected by at least one such distortion type and to locate areas which necessarily contain shadows can be cumbersome. Hence, we require a method which can detect shadows with little computational overhead and takes into consideration the complicated nature of SAR image acquisition process.

In this letter we propose a simple approach based on $1$-$D$ geometrical projections using the digital elevation model (DEM) corresponding to the geo-referenced SAR image. We first rotate and fit the DEM to the radar geometry so that each row would be in the line of sight of the radar. Then the shadow extraction is done row by row according the image size, thus simplifying the process to 1-dimensional space. The proposed approach can provide better target classification and a precise height estimation by using only the shadow information in high resolution SAR images. 

The rest of the paper is presented as follows. In Section~\ref{sec:shadow} we start with the characteristics of SAR imagery and then we illustrate our methodology for shadow detection. Experimental results are given next in Section~\ref{sec:exper}. Finally, Section~\ref{sec:concl} concludes the paper. A preliminary version appeared in~\cite{Haddad}.

\section{Shadow characterization and detection in SAR images}\label{sec:shadow}
\begin{figure}
	\centering
	\includegraphics[width=10cm, height=6cm]{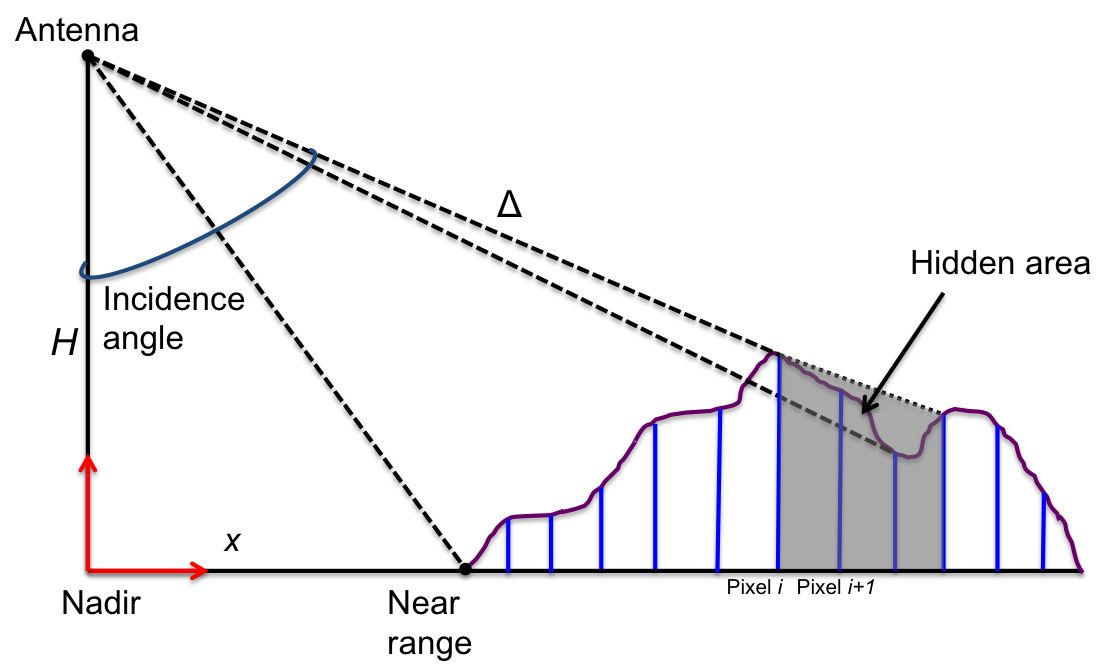}
	\caption{llustration of shadow phenomenon on a given row of SAR image.}\label{fig:angle}
\end{figure}
Detecting radar shadows includes different characterizations and several different definitions give rise to different detection methods~\cite{Mikhail}. Here we provide factors which affect the shading are introduced and a modeling mechanism is introduced. The incidence angle (Figure~\ref{fig:angle}) is defined as the angular shift from the vertical position of the sensor viewing in the direction of the target. It is an important parameter which manages the shape of the shadow. For a fixed distance from the nadir, the shadow is an increasing function of the incidence angle. If we increase the sensor height with respect to a given target then the shadow regions shrinks~\cite{Hansen, Zhang}. However, the resolution of the target gets reduced as we move away from the target. On the other hand for a fixed altitude of the sensor, there are other parameters which can affect the amount shadow regions, for example the target altitude can influence the shadow regions. If the slope of the overall terrain is higher than the depressing angle, then the shaded areas return very weak responses to the sensor. Moreover, the regions with high relief are due to low depression angle and is a source of large shadows present. Thus, we see that the loss of information due to shadows is a decreasing function of the depression angle. Radar shadows produce a $3$D visualization effect without using a stereoscope, but the information provided by the shadow effect on the altitude is not reliable. Note that a radar image superimposed with shadows is not show significant elevations close to nadir. Moreover, the shadows closer to nadir are not detected at all.

\begin{figure}
	\centering
	\subfigure[Map]{\includegraphics[width=4.5cm, height=3.75cm]{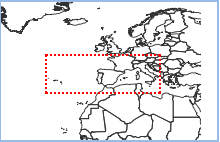}}
	\subfigure[ROI Selection]{\includegraphics[width=4.5cm, height=3.75cm]{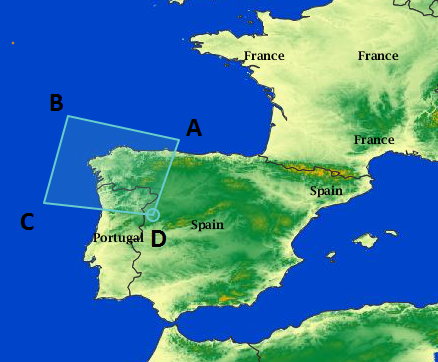}}
	\subfigure[Coordinates]{\includegraphics[width=4.5cm, height=3.75cm]{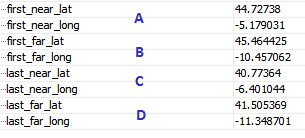}}\\
	
	\subfigure[Extracted Region]{\includegraphics[width=4.5cm, height=3.5cm]{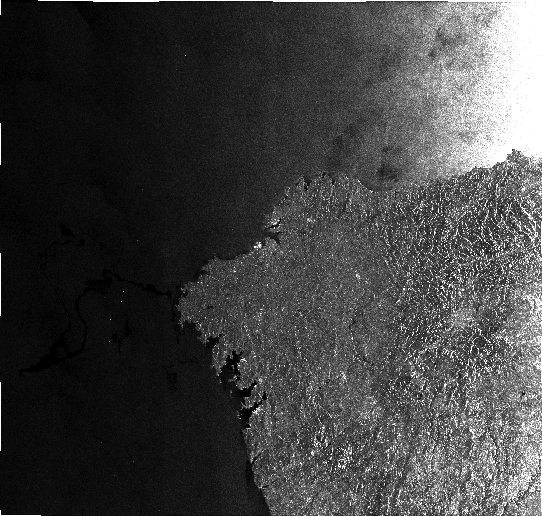}}
	\subfigure[Reprojected]{\includegraphics[width=4.5cm, height=3.5cm]{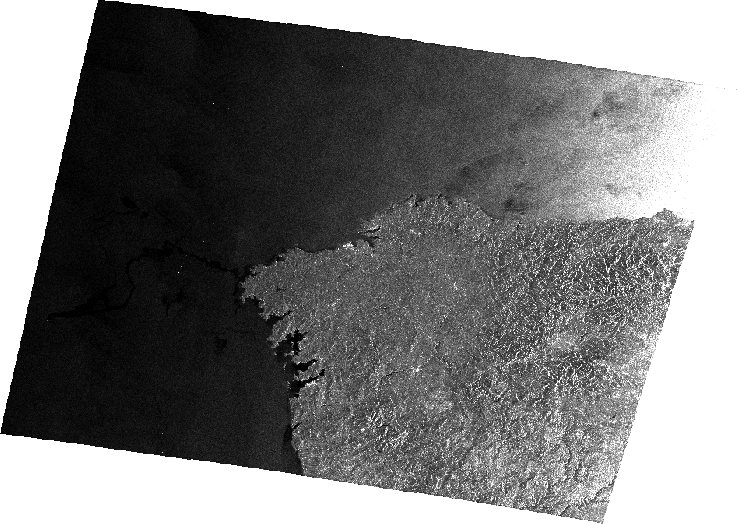}}
	\subfigure[Incidence Angle]{\includegraphics[width=4.5cm, height=3.5cm]{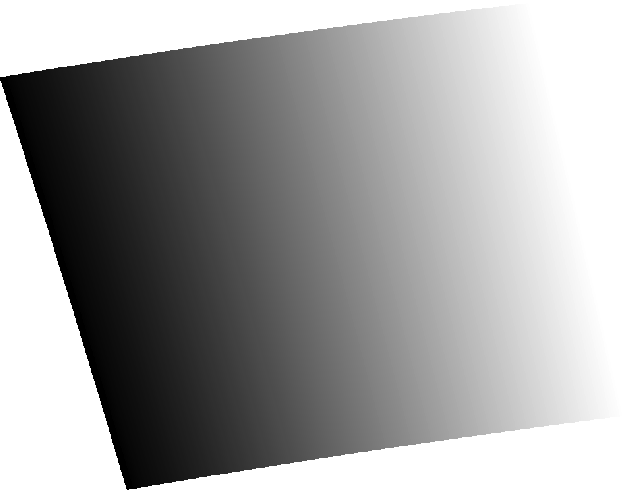}}
	\caption{llustration of the region/DEM region of interest (ROI) extraction from the north of Portugal and Spain from USGS. 
	(a) The region is shown in a global map for illustration purpose.
	(b) The rectangular selection region with ABCD indicating the corner points is selected.
	(c) Co-ordinates of the four corner regions.
	(d) Extracted ROI.
	(e) Reprojected/geo-referenced image.
	(f) Incidence Angle map.}\label{fig:usgs}
\end{figure}

\begin{figure}
	\centering
	\includegraphics[width=14cm, height=8.5cm]{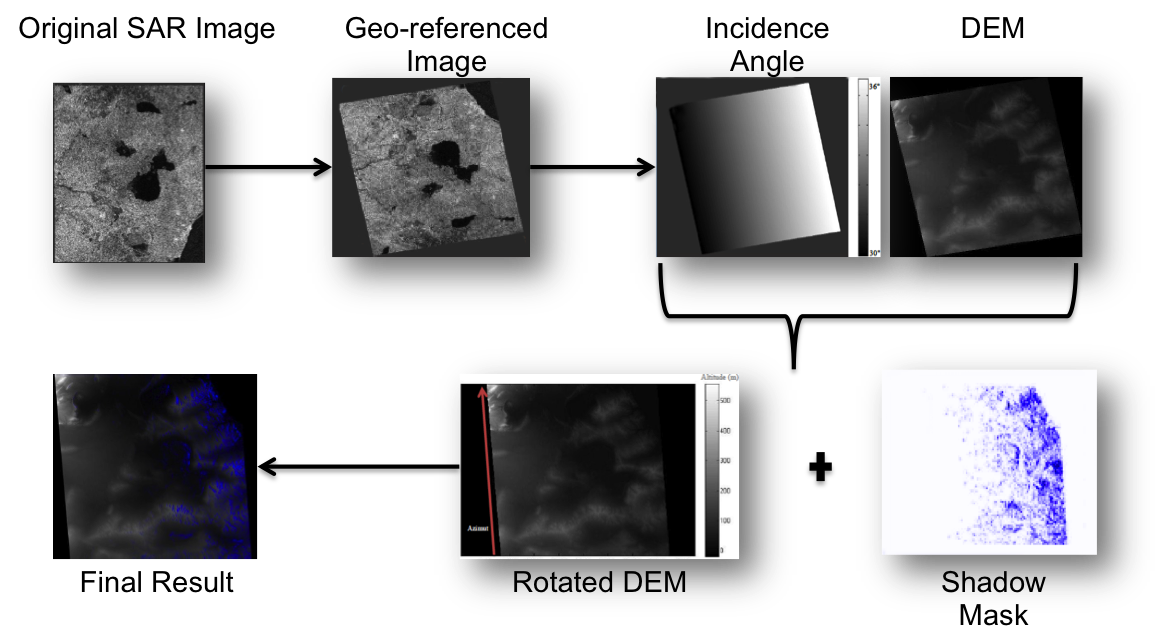}
	\caption{Shadow detection algorithm. The flowchart corresponds to the DEM based $1$D projections scheme and final result shows the shadow regions in blue color. The azimuth is shown as a red arrow in the rotated DEM image. For color referred in this figure see the online PDF version.}\label{fig:flow}
\end{figure}

Based on the above observations and the fact that SAR recovers a given scene line by line we propose the following steps for effective identification of shadow regions.
\begin{enumerate}
\item The given SAR image is first geo-referenced. For this purpose, we utilize the BEAM (Basic ENVISAT Toolbox
for (A)ATSR and MERIS) software distributed by the European Space Agency (ESA) for SAR image geo-referencing as we tested the developed approach on ASAR ENVISAT data primarily.

\item For each radar line of sight with respect to the antenna height we generate an incidence angle image which is obtained by row by row consideration.

\item The DEM is rotated so that each row would match with that of the radar line of sight (slant range). For validation we use a $30m$ resolution ASTER DEM\footnote{\href{http://asterweb.jpl.nasa.gov/gdem.asp}{http://asterweb.jpl.nasa.gov/gdem.asp}} downloaded from the USGS (United States Geological Survey) website\footnote{\href{http://gdex.cr.usgs.gov/gdex/}{http://gdex.cr.usgs.gov/gdex/}}, see Figure~\ref{fig:usgs} for example.

\item The shadow mask is generated from the rotated DEM using the $1$D geometrical projections applied to each row of the DEM. That is, if we let $H$ be the sensor height, $t$ the number of pixels, then the projection line is given by, 
\begin{eqnarray}\label{E:1dproj}
l(x)  =a(x)\cdot x+ H,
\end{eqnarray}
where $a(x)$ represents the slope value computed at each pixel $x$ in the line of sight, see Figure~\ref{fig:angle}.
Note that $H$ is the tunable parameter indicating the height where the antenna is present.
Then, a superposition of the DEM and the shadow mask is done, see Figure~\ref{fig:flow} the blue color regions.
\end{enumerate}

Thus, the methodology we proposed composed of two major parts. First, the shadow detection on the DEM profile is done in a row by row fashion. Second the DEM and shadow mask are merged together to obtain the result. Figure~\ref{fig:flow} summarizes the proposed approach. The flowchart given in Figure~\ref{fig:flow} explains all the five steps previously mentioned steps starting with a given SAR image with corresponding DEM. We note that in our study the data elevation model is presented in a grayscale image from zero meter (Black) to $250 m$ (White), also the Incidence Angle is presented in a grayscale image from 30$^\circ$ (Black) to 36$^\circ$ (White).
\section{Experimental results}\label{sec:exper}
\begin{figure}
	\centering
		\subfigure[1D Signals]{\includegraphics[width=3.3cm, height=3cm]{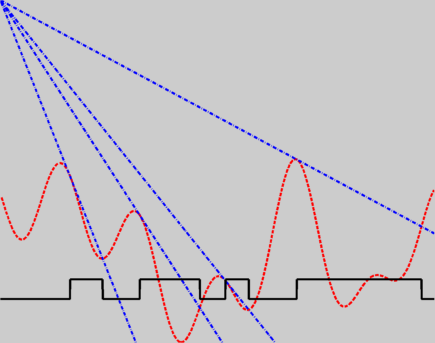}
		\includegraphics[width=3.3cm, height=3cm]{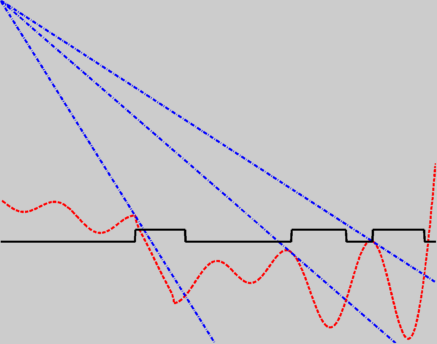}
		\includegraphics[width=3.3cm, height=3cm]{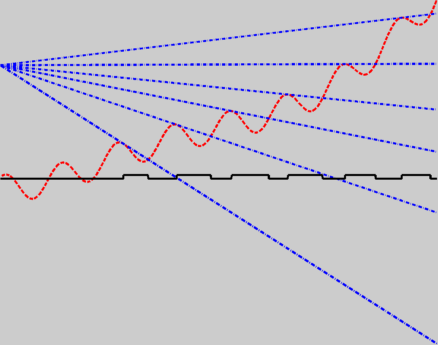}
		\includegraphics[width=3.3cm, height=3cm]{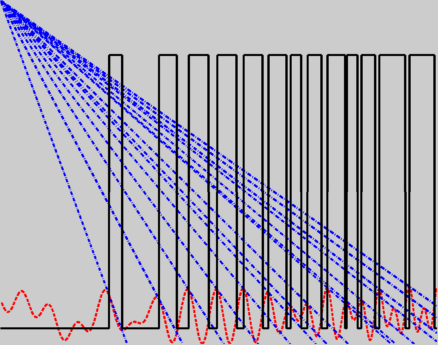}}\\
		
		\subfigure[Original1]{\includegraphics[width=3.3cm, height=3cm]{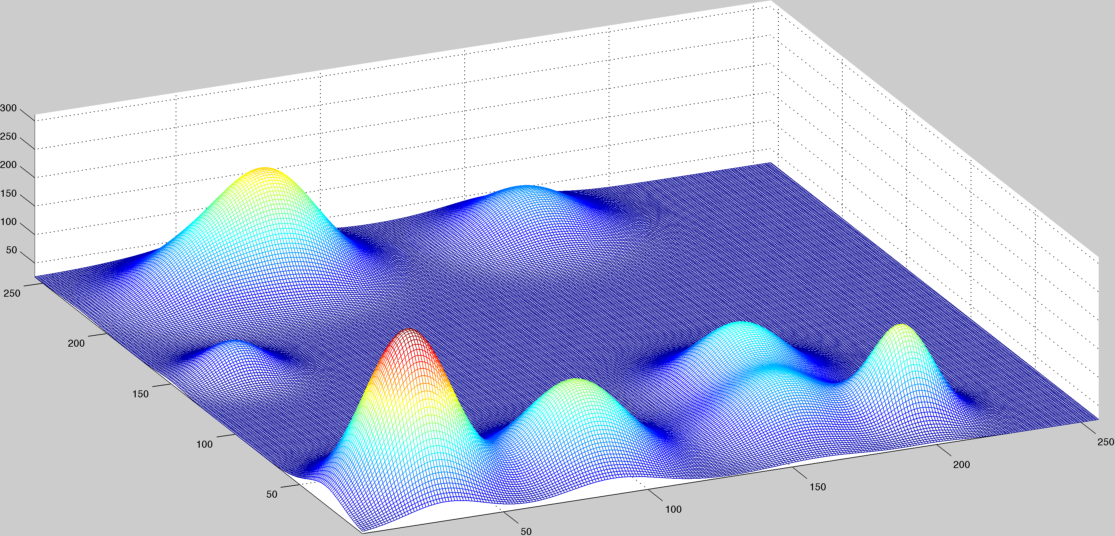}}
		\subfigure[Original2]{\includegraphics[width=3.3cm, height=3cm]{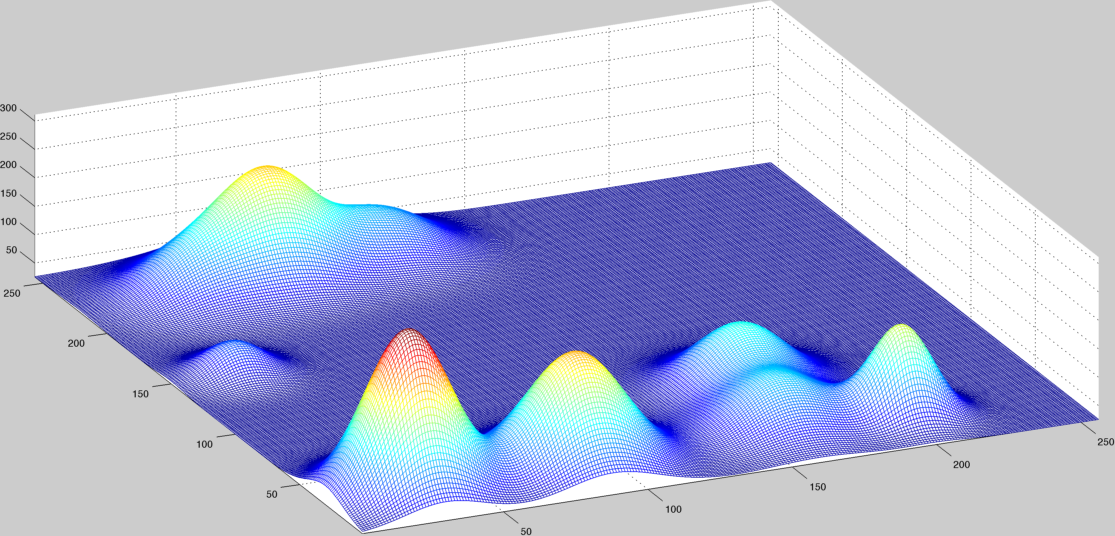}}
		\subfigure[Original3]{\includegraphics[width=3.3cm, height=3cm]{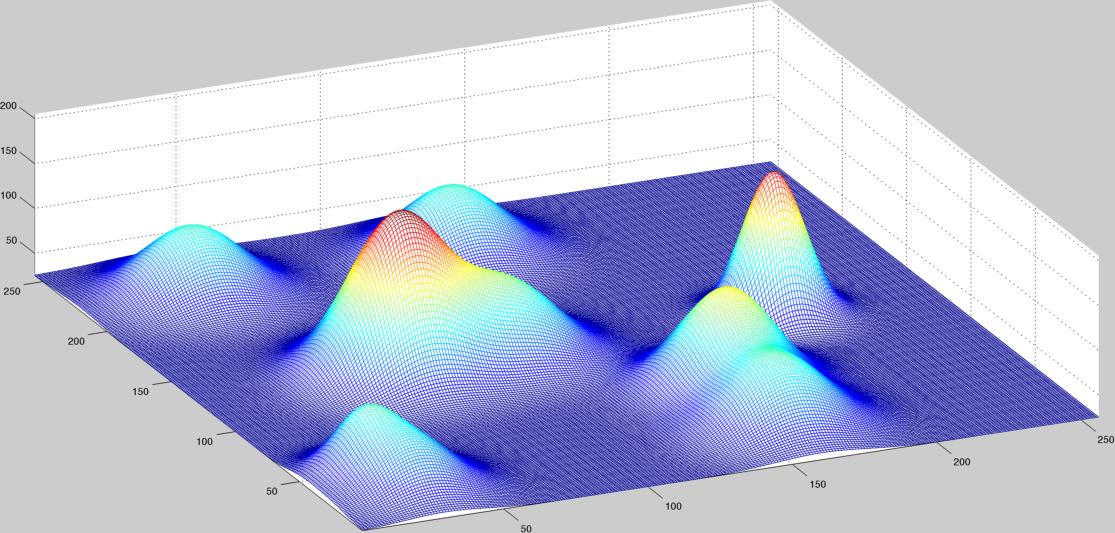}
		\includegraphics[width=3.3cm, height=3cm]{SyntheticGaussian_Original3_3D.png}}\\

		\subfigure[Same Height $H=400m$]{\includegraphics[width=3.3cm, height=3cm]{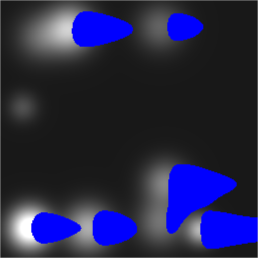}\,
		\includegraphics[width=3.3cm, height=3cm]{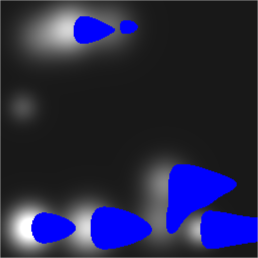}}\,
		\subfigure[Height $H=400m$ (left), Height $H=200m$ (right)]{\includegraphics[width=3.3cm, height=3cm]{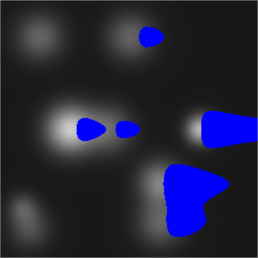}
		\includegraphics[width=3.3cm, height=3cm]{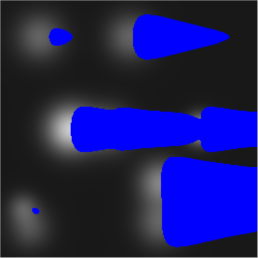}}

		\subfigure[2D Shadow Detection]{\includegraphics[width=3.3cm, height=3cm]{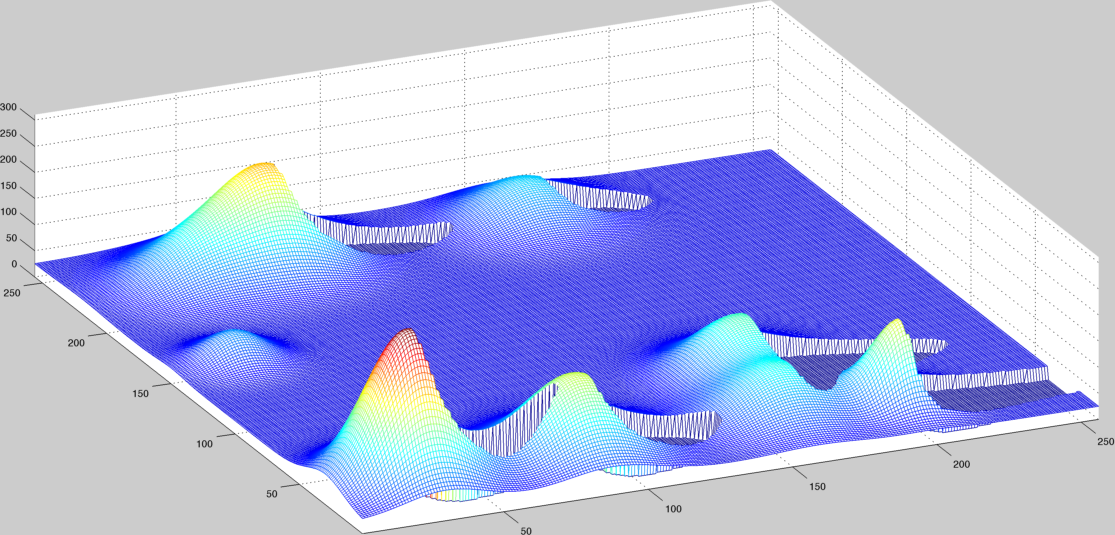}\,
		\includegraphics[width=3.3cm, height=3cm]{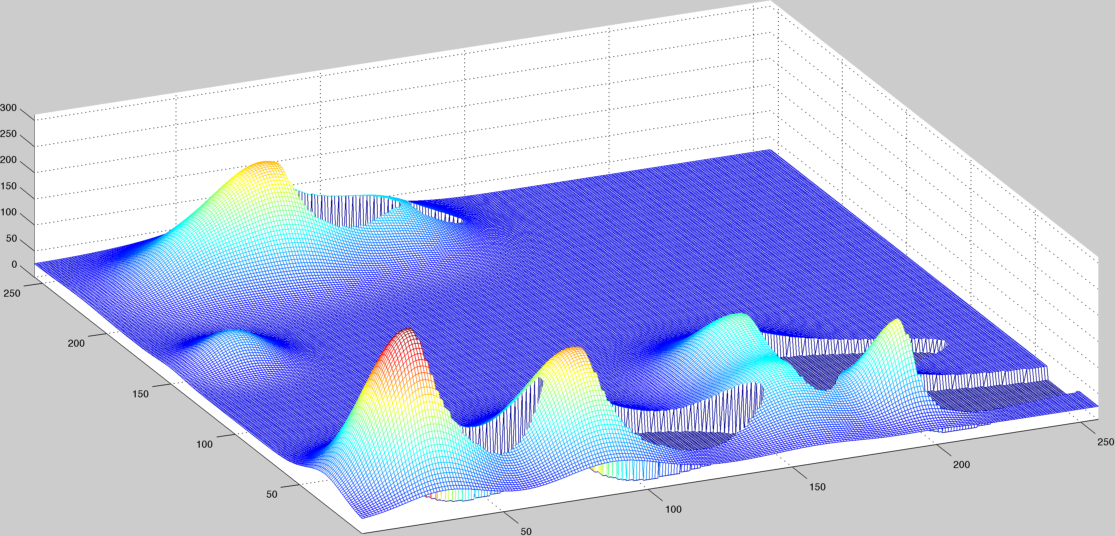}\,
		\includegraphics[width=3.3cm, height=3cm]{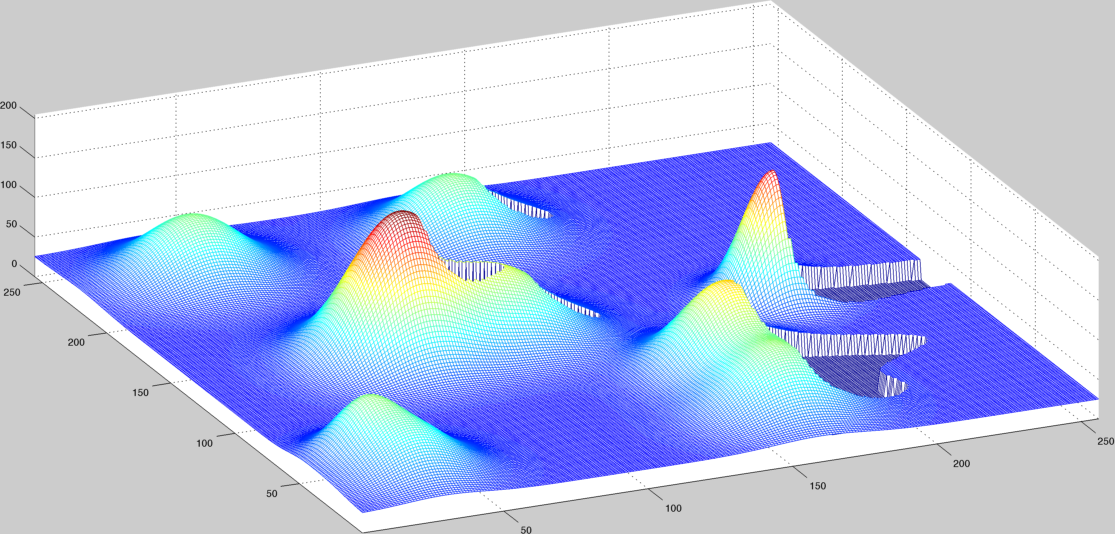}
		\includegraphics[width=3.3cm, height=3cm]{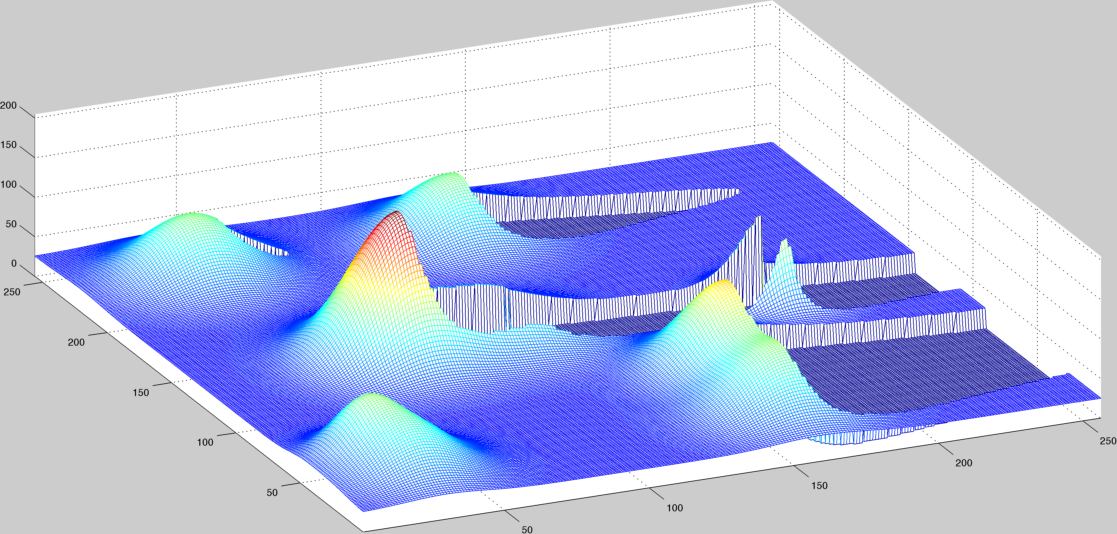}}
	\caption{Synthetic test images illustrating shadow detection method. 
	(a) $1$D signals generated using sinusoidal, logarithmic and Gaussian functions respectively. 
	The blue ( slash-dotted, $-.-$) lines are the $1$D projections, the red (slash, $---$) is the DEM or the terrain and the black solid lines are the detected shadows. 
	We use multiple $2$D Gaussian functions modeling various relief scenarios in surface visualisation (b, c) different mountains + same Radar Height, (d) Same mountains + different Radar Heights.
	(e)-(f) Corresponding detected shadow regions for (b,c)-(d). 
	(g) Corresponding surface visualization with shadows superimposed on the terrain.}\label{fig:synth}
\end{figure}

\begin{figure}
	\centering
	\includegraphics[width=3.25cm, height=3cm]{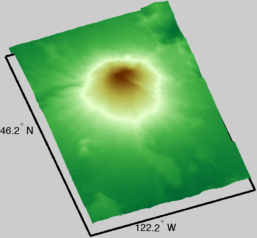}
	\includegraphics[width=3.25cm, height=3cm]{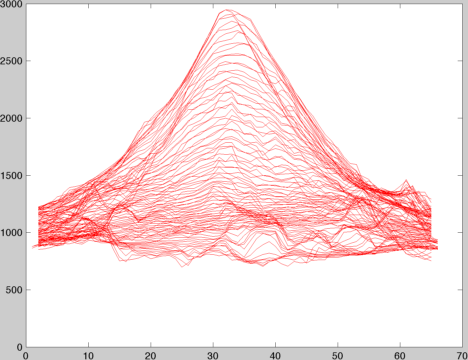}
	\includegraphics[width=3.25cm, height=3cm]{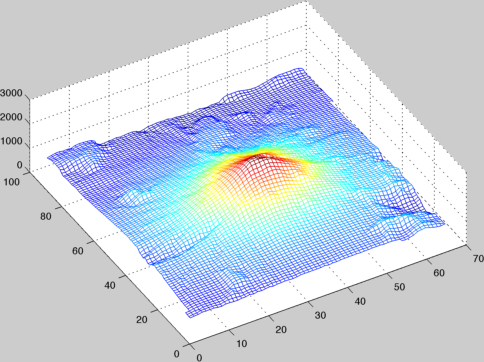}
	\includegraphics[width=2.25cm, height=3cm]{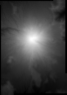}
	
	\subfigure[DEM]{\includegraphics[width=3.25cm, height=3cm]{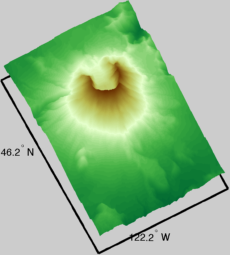}}
	\subfigure[Projections]{\includegraphics[width=3.25cm, height=3cm]{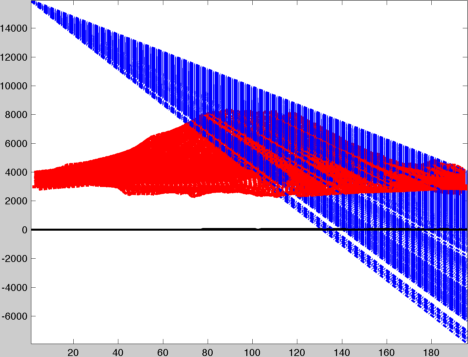}}
	\subfigure[Shadows]{\includegraphics[width=3.25cm, height=3cm]{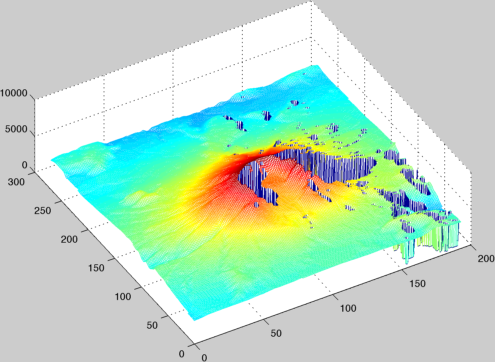}}
	\subfigure[2D]{\includegraphics[width=2.25cm, height=3cm]{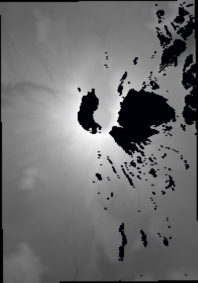}}
	
	\caption{Example DEM areas and corresponding shadow detection results.
	(a) DEM
	(b) 1D projections
	(c) Detected shadows using the DEM
	(d) Detected shadows projected onto the 2D image for visualization. First row the no shadows case (size $94\times66$) and the second row shadows present case (size $280\times196$).}\label{fig:demo}
\end{figure}

To test the robustness of the proposed shadow detection approach we perform some simulated profiles corresponding to different 1D signals using sinusoidal, logarithmic and Gaussian functions. The final detection results shown in Figure~\ref{fig:synth}(a), clearly indicate the shadow regions detected as it consist of finding the intersection of the radar line with the elevated profile from the DEM.  Here the shadow regions are indicated as $1$D square signal (a shadow mask~\cite{Rees}, in black color) where the zero (flat) values indicate the absence of shadows and non-zero (bumps) values indicate detected shadow intervals. We next use multiple $2$D Gaussian functions to simulate different relief levels in an image. To reiterate our approach, we consider the following steps:
\begin{enumerate}
\item[1.] Divide the initial model line by line.
\item[2.] Shadow detection for each line of sight using projections along the lines.
\item[3.] Superposition of the shadow mask on the Gaussian DEM.
\end{enumerate}
Figure~\ref{fig:synth} (b-d) show synthetic $2$D Gaussian functions simulating various scenarios and corresponding shadow regions detected by our approach are provided in 2D image regions (blue, Figure~\ref{fig:synth} (e-h)), and in surface visualisation (Figure~\ref{fig:synth} (g)) . It is clear that we do not detect shadows near the nadir. Nevertheless, the detected shadow region's shape provides a confirmation of variations in the relief. Figure~\ref{fig:demo} show two real  examples for the shadow detection algorithm with respect to different DEMs (Figure~\ref{fig:demo} (a)) of smaller dimensions and corresponding results with the same radar height $16~km$ in (Figure~\ref{fig:demo} (c-d)). This illustrates the robustness of the 1D projections (Figure~\ref{fig:demo} (b)) based scheme. 

\begin{figure}
	\centering
	\subfigure[DEM]{\includegraphics[width=5cm, height=3.5cm]{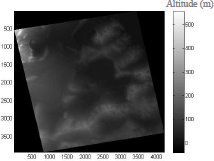}}
	\subfigure[$5~km$]{\includegraphics[width=4.25cm, height=3.25cm]{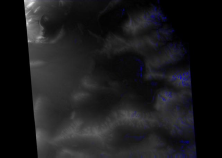}}
	\subfigure[$3~km$]{\includegraphics[width=4.25cm, height=3.25cm]{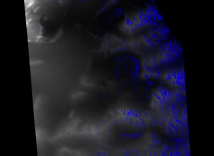}}

	\subfigure[DEM]{\includegraphics[width=5cm, height=3.75cm]{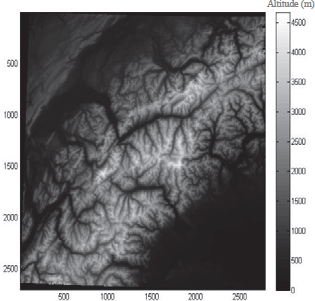}}
	\subfigure[$50~km$]{\includegraphics[width=4.25cm, height=3.75cm]{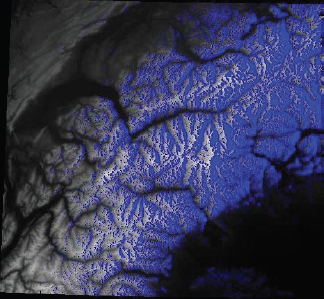}}
	\subfigure[$150~km$]{\includegraphics[width=4.25cm, height=3.75cm]{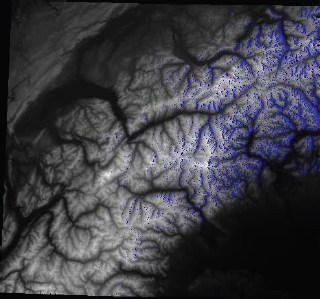}}
	
	\caption{Shadow detection results for two regions with variation in reliefs and with different altitudes. 
	Top row: Tunisia. 
	Bottom row: Alps mountains. Different altitudes with detected shadow regions are given.}\label{fig:real}
\end{figure}

Next in Figure~\ref{fig:real} we consider SAR images and their corresponding DEMs to test our algorithm further.  Figure~\ref{fig:real} (a)\&(d) show two different DEM regions with different characteristics. The final shadow detection results for different radar heights are given in Figure~\ref{fig:real} (b-c)\& (e-f) respectively. As can be seen the scheme provides reliable shadow estimation across the variation in reliefs. 

Further visualization of the detected shadow regions and corresponding DEMs are available online~\footnote{\href{http://dx.doi.org/10.6084/m9.figshare.659896}{http://dx.doi.org/10.6084/m9.figshare.659896}}where MATLAB figure files showing $3$D maps visualisations and data files (\textit{.fig, .mat}) are provided as well~\footnote{\href{http://sites.google.com/site/suryaiit/research/shadows}{http://sites.google.com/site/suryaiit/research/shadows}}.  We believe that the approach presented here can augment colored remote sensed data based approach~\cite{Liu} for the recognition of backscattering objects on SAR images via their shadows.
\section{Conclusion}\label{sec:concl}
In radar imaging the shadows can give additional information which can be used for target recognition. For example, the target height can be obtained by the geometry of observation system and the length of shadow, assuming either level or known ground~\cite{Liu,Soergel}. In this paper a simple solution is given to identify the shadow regions using DEM and $1$D projections. Experimental results with synthetic and real DEMs indicate the robustness of our approach in obtaining  shadow regions. 

\bibliographystyle{plain}
\bibliography{SARshadowsrefs} 

\begin{thebibliography}{10}

\bibitem{AbramsSfShadow13}
A.~Abrams, K.~Miskell, and R.~Pless.
\newblock The episolar constraint: Monocular shape from shadow correspondence.
\newblock In {\em CVPR}. IEEE, 2013.

\bibitem{AdelineVHRshadow13}
K.~R.~M. Adeline, M.~Chen, X.~Briottet, S.~K. Pang, and N.~Paparoditis.
\newblock Shadow detection in very high spatial resolution aerial images: A
  comparative study.
\newblock {\em ISPRS Journal of Photogrammetry and Remote Sensing},
  80(11):21--38, 2013.

\bibitem{ComberHRshadow12}
A.~Comber, M.~Umezaki, R.~Zhou, Y.~Ding, Y.~Li, H.~Fu, H.~Jiang, and
  A.~Tewkesbury.
\newblock Using shadows in high-resolution imagery to determine building
  height.
\newblock {\em Remote Sensing Letters}, 3(7):551--556, 2012.

\bibitem{Haddad}
O.~Haddad, R.~Abdelfattah, and H.~Ajili.
\newblock Extracting radar shadow from {SAR} images.
\newblock In {\em IGARSS}, pages 2101--2104. IEEE, 2012.

\bibitem{Hansen}
R.~E. Hansen, H.~J. Callow, and J.~Groent.
\newblock Enhancing target shadow in {SAR} images.
\newblock {\em Electronics letters}, 43(5):69--70, 2007.

\bibitem{Jahangir}
M.~Jahangir, D.~Blacknell, C.~P. Moate, and R.~D. Hill.
\newblock Extracting information from shadows in {SAR} imagery.
\newblock In {\em International Conference on Machine Vision (ICMV)}, pages
  107--112. IEEE, 2007.

\bibitem{Liu}
J.~Liu, T.~Fang, and D.~Li.
\newblock Shadow detection for color remotely sensed images based on
  multi-feature integration.
\newblock {\em SPIE Journal of Remote Sensing}, 6, 2012.

\bibitem{Mikhail}
E.~Mikhail, J.~Bethel, and J.~McGlone.
\newblock {\em Introduction to modern photogrammetry}.
\newblock John Wiley and Sons, 2001.

\bibitem{Papson}
S.~Papson.
\newblock Modeling of target shadows for {SAR} image classification.
\newblock In {\em IEEE Applied Imagery and Pattern Recognition Workshop
  (AIPR)}, 2006.

\bibitem{Raney}
R.~K. Raney.
\newblock {\em Manual of Remote Sensing}, volume~2, chapter Radar fundamentals:
  Technical perspective, pages 9--130.
\newblock John Wiley and Sons, New York, 3rd edition, 1998.

\bibitem{Rees}
W.~G. Rees.
\newblock Simple masks for shadowing and highlighting in {SAR} images.
\newblock {\em International Journal of Remote Sensing}, 21(11):2145--2152,
  2000.

\bibitem{SaninShadowSurveyPR12}
A.~Sanin, C.~Sanderson, and B.~C. Lovell.
\newblock Shadow detection: A survey and comparative evaluation of recent
  methods.
\newblock {\em Pattern Recognition}, 45(4):1684--1695, 2000.

\bibitem{Soergel}
U.~Soergel, U.~Thoennessen, and U.~Stilla.
\newblock Visibility analysis of man-made objects in {SAR} images.
\newblock In {\em GRSS/ISPRS Joint Workshop on Remote Sensing and Data Fusion
  over Urban Areas (DFUA)}, 2003.

\bibitem{TongLinHRshadows13}
X.~Tong, X.~Lin, T.~Feng, H.~Xie, S.~Liu, Z.~Hong, and P.~Chen.
\newblock Use of shadows for detection of earthquake-induced collapsed
  buildings in high-resolution satellite imagery.
\newblock {\em ISPRS Journal of Photogrammetry and Remote Sensing}, 79:53--67,
  2013.

\bibitem{Zhang}
Z.~B. Zhang.
\newblock Alternative focus range estimation method for fixed focus shadow
  enhancement in {SAR} images.
\newblock {\em International Journal of Remote Sensing}, 47(25):1393--1394,
  2011.

\end{thebibliography}
\end{document}